\documentclass[journal,10pt]{IEEEtran}
\ifCLASSINFOpdf
\else
\fi
\usepackage{comment}

\usepackage{epsfig}
\usepackage{graphicx}
\usepackage{amsmath}
\usepackage{amssymb}

 \usepackage{algorithmic}
 
 \usepackage{cases}

\usepackage[linesnumbered,ruled,vlined]{algorithm2e}

\usepackage{amsmath,graphicx}

\usepackage{pstricks,graphicx}
\usepackage{setspace}
\usepackage{psfrag}

\usepackage{multirow}

\usepackage{amsmath,graphicx}

\usepackage{pstricks,graphicx}
\usepackage{setspace}
\usepackage{psfrag}
\usepackage{amssymb}
\usepackage{amsmath}
\usepackage{amsthm}

\newtheorem{remark}{Remark}

\newtheorem{lemma}{Lemma}

\newtheorem{theorem}{Theorem}

\newtheorem{definition}{Definition}

\begin{document}



\title{Extending AdamW by Leveraging Its Second  Moment and Magnitude}

\graphicspath{{figures/}}
%

\author{Guoqiang~Zhang,  Kenta Niwa and W. Bastiaan Kleijn
\thanks{G.~Zhang is with the School of Electrical and Data Engineering, University of Technology, Sydney, Australia. Email: {guoqiang.zhang@uts.edu.au}}
\thanks{K.~Niwa is with Nippon Telegraph and Telephone (NTT) Corporation, Japan.
Email: {kenta.niwa.bk@hco.ntt.co.jp}}
\thanks{W. Bastiaan Kleijn is with the School of Engineering and Computer Science, Victoria University of Wellington, New Zealand. Email: {bastiaan.kleijn@ecs.vuw.ac.nz}}
}

\maketitle

\begin{abstract}
Recent work  \cite{Bock19Adam} analyses the local convergence of Adam in a neighbourhood of an optimal solution for a twice-differentiable function.  It is found that the learning rate has to be sufficiently small to ensure local stability of the optimal solution. The above convergence results also hold for AdamW. In this work, 
we propose a new adaptive optimisation method by extending AdamW in two aspects with the purpose to relax the requirement on small learning rate for local stability, which we refer to as Aida.  Firstly, we consider tracking the 2nd moment $r_t$ of the $p$th power of the gradient-magnitudes. $r_t$ reduces to $v_t$  of AdamW when $p=2$. Suppose $\{m_{t}\}$ is the first moment of AdamW. It is known that the update direction $m_{t+1}/\sqrt{v_{t+1}+\epsilon}$  (or $m_{t+1}/\sqrt{v_{t+1}}+\epsilon$) of AdamW (or Adam) can be decomposed as the sign vector $\textrm{sign}(m_{t+1})$ multiplied elementwise by a vector of magnitudes $|m_{t+1}|/(\sqrt{v_{t+1}+\epsilon})$ (or $|m_{t+1}|/(\sqrt{v_{t+1}}+\epsilon)$).  Aida is designed to compute the $q$th power of the magnitude in the form of $|m_{t+1}|^q/({r_{t+1}}+\epsilon)^{q/p}$ (or $|m_{t+1}|^q/({r_{t+1}}^{q/p}+\epsilon)$), which reduces to that of AdamW when $(p,q)=(2,1)$.   

Suppose the origin $0$ is a local optimal solution of a twice-differentiable function. It is found theoretically that when $q>1$ and $p>1$ in Aida,  the origin $0$ is locally stable only when the weight-decay is non-zero. 
Experiments are conducted for solving ten toy optimisation problems and training Transformer and Swin-Transformer for two deep learning (DL) tasks. The empirical study demonstrates that in a number of scenarios (including the two DL tasks), Aida with particular setups of $(p,q)\neq (2,1)$ outperforms the setup $(p,q)=(2,1)$ of AdamW.    

\end{abstract}



\vspace{-2mm}
\section{Introduction}
\vspace{-1mm}
\label{sec:Intro}

Stochastic gradient descent (SGD) and its variants have been widely applied in deep learning due to their simplicity and effectiveness \cite{Lecun15nature}.  In the literature, SGD with momentum (also known as heavy ball (HB) method \cite{Sutskever13NAG, Polyak64CM}) dominates other optimisers for image classification tasks \cite{He15ResNet, Wilson17AdamNegative}. Suppose the objective function $f({x}):x\in \mathbb{R}^d$ is differentiable. The update expression of HB for minimising $f(x)$ can be represented as
\begin{align}
{m}_{t+1} & = \beta_t {m}_{t} + \alpha_{t} g(x_t)  \label{equ:SGDM1}\\
{x}_{t+1} &= {x}_{t} - \gamma_t {m}_{t+1},  \label{equ:SGDM2}
\end{align}
where $g(x_t) = \nabla f({x}_{t})$ is the gradient at ${x}_{t}$,  $\beta_t$ is the averaging coefficient of the momentum,  and $\alpha_{t}$ with fixed $\gamma_t=1$ or $\gamma_t$ with fixed  $\alpha_t=1$ is the common learning rate for all the coordinates of ${x}$. In practice, the HB method is often combined with a certain learning-rate scheduling strategy when training DNN models.  



\begin{algorithm}[t]
   \caption{\small Aida for optimisation of $f({x})$}
   \label{alg:Aida}
\begin{algorithmic}[1]
   \STATE {\small {\bfseries Input:} $\beta_1$, $\beta_2$,  $p\geq 1$, $q\geq 1$, $\eta$, $\epsilon \geq 0$, $\mu\geq 0$ }
   \STATE {\small {\bfseries  Init.:} ${x}_0\in \mathbb{R}^d$,  ${m}_0 = 0$, ${r}_0 = 0$ }
   \FOR{\small $t=0, 2, \ldots, t_1$}
   \STATE \hspace{-2mm}{\small  ${g}_t \leftarrow \nabla f({x}_{t}) $  }
   \STATE \hspace{-2mm}{\small ${m}_{t+1} \leftarrow (1-\beta_1) {m}_{t}  + \beta_1 {g}_t$ }
   \STATE \hspace{-2mm}{\small ${r}_{t+1} \leftarrow (1-\beta_2) {r}_{t}  + \beta_2 |{g}_t|^p $  }
   \STATE \hspace{-2mm}{\small $  {x}_{t+1} \hspace{-0.6mm}\leftarrow \hspace{-0.6mm} (1\hspace{-0.6mm}-\hspace{-0.6mm}\mu){x}_{t} \hspace{-0.6mm} - \hspace{-0.6mm}  \left\{\hspace{-2.5mm} \begin{array}{l}  \eta \frac{ (1-\beta_2^{t+1}))^{q/p} }{(1-\beta_1^{t+1})^q}  \frac{|{m}_{t+1}|^{q}\odot \textrm{sign}(m_{t+1}) }{(r_{t+1}+\epsilon)^{q/p}}  \\
  \eta \frac{ (1-\beta_2^{t+1}))^{q/p} }{(1-\beta_1^{t+1})^q}  \frac{|m_{t+1}|^{q}\odot \textrm{sign}(m_{t+1}) }{(r_{t+1})^{q/p}+ \epsilon} 
    \end{array} \right. $ }    
   \ENDFOR 
   \STATE {\bfseries Output:} {\small ${x}_{t_1}$  }
   \\ 
  
  \vspace{1mm}\hrule  width0.45\textwidth \vspace{1mm}
 \hspace{-5mm} *\hspace{0.1mm} {\small {Remark}: Aida reduces to AdamW when $p=2$ and $q=1$. } 
\end{algorithmic}
\end{algorithm}


To bring flexibility to the HB method, one active research trend in last decade is to introduce individual learning rates for all the components of ${m}_{t+1}$ in (\ref{equ:SGDM2}), referred to as \emph{adaptive optimisation}. Duchi et. al \cite{Duchi11AdaGrad}, were the first to track the moment of squared-gradients in computation of the individual learning rates. 
The basic principle is to identify and multiply those coordinates of $m_{t+1}$ that have historically small squared-gradients on average with relatively large individual learning rates to enhance their contributions to the updates of $x_{t+1}$, and vice versa. 
Following the work of \cite{Duchi11AdaGrad}, various adaptive optimisation methods have been proposed for computing more effective individual learning rates under different scenerios. The methods include, for example, RMSprop \cite{Tieleman12RMSProp}, Adam \cite{Kingma17},  NAdam \cite{Dozat16NAdam}, AMSGrad \cite{Reddi18Amsgrad}, Adafactor \cite{Shazeer18Adafactor},  AdamW \cite{Loshchilov19AdamW}, and LAMB\cite{You20Lamb}. 

In the literature,  Adam and its variant AdamW are probably most popular among the existing adaptive optimisation methods (e.g., \cite{Transformer17, liu2021Swin, Zhang2019Adam, Devlin18Bert}). The update expression of AdamW at iteration $t$ can be written as
\begin{align}
\hspace{0mm}{m}_{t+1} & =\beta_1{m}_{t} + (1-\beta_1) g(x_t)      \qquad \qquad  \qquad  \quad \label{equ:Adam1}\\
\hspace{0mm}{v}_{t+1} & = \beta_2 {v}_{t} + (1-\beta_2) |g(x_t)|^2  \qquad  \qquad \qquad \quad \label{equ:Adam2} 
\end{align}
\vspace{-5mm}
\begin{subnumcases}{\hspace{-5mm}\textrm{or}  \hspace{-0.8mm}}
\hspace{-2mm}  {x}_{t\hspace{-0.2mm}+\hspace{-0.2mm}1}    \hspace{-0.8mm} = \hspace{-0.8mm}  (1 \hspace{-0.8mm}- \hspace{-0.8mm}\mu){x}_{t}  \hspace{-0.8mm} -  \hspace{-0.8mm} \eta \frac{\sqrt{1  \hspace{-0.8mm} - \hspace{-0.8mm}   \beta_2^{t\hspace{-0.2mm}+\hspace{-0.2mm}1}}}{1-\beta_1^{t+1}} \frac{|{m}_{t\hspace{-0.2mm}+\hspace{-0.2mm}1}| \hspace{-0.8mm}\odot \hspace{-0.8mm} \textrm{sign}(m_{t\hspace{-0.2mm}+\hspace{-0.2mm}1}) }{\sqrt{{v}_{t+1}+\epsilon}} \label{equ:Adam3_1} \\
\hspace{-2mm} {x}_{t\hspace{-0.2mm}+\hspace{-0.2mm}1}    \hspace{-0.8mm} = \hspace{-0.8mm}  (1 \hspace{-0.8mm}- \hspace{-0.8mm}\mu){x}_{t}   \hspace{-0.8mm} - \hspace{-0.8mm}  \eta \frac{\sqrt{1  \hspace{-0.8mm} - \hspace{-0.8mm}    \beta_2^{t\hspace{-0.2mm}+\hspace{-0.2mm}1}}}{1-\beta_1^{t+1}}  \frac{|m_{t\hspace{-0.2mm}+\hspace{-0.2mm}1}| \hspace{-0.8mm}\odot\hspace{-0.8mm}  \textrm{sign}(m_{t\hspace{-0.2mm}+\hspace{-0.2mm}1}) }{\sqrt{{v}_{t+1}}+\epsilon} \label{equ:Adam3_2} 
\end{subnumcases}
where $g(x_t)=\nabla f({x}_{t})$, $\odot$ and $/$ denote element-wise multiplication/division, and $0<\beta_1, \beta_2<1$,  $\epsilon>0$, $1>\mu\geq 0$. The parameter $\eta$ is the common learning rate. $m_t$ and $v_t$ are the two moments of gradients and squared-gradients over iterations, respectively. The parameter  $\mu$ governs a weight decay and $\mu=0$ leads to the update expression of Adam. Note that the update for $x_{t+1}$ may have different forms of expression depending on how the parameter $\epsilon$ participates in computation of the individual learning rates. As indicated by (\ref{equ:Adam3_1}) for example, the update direction $\frac{|{m}_{t+1}|\odot \textrm{sign}(m_{t+1}) }{\sqrt{{v}_{t+1}+\epsilon}}$ for $x_{t+1}$ can be viewed as the sign vector $\textrm{sign}(m_{t+1})$ multiplied element-wise by the magnitude vector $\frac{|{m}_{t+1}|}{\sqrt{{v}_{t+1}+\epsilon}}$.

Several empirical studies have been conducted attempting to understand the reason behind the effectiveness of Adam.  For instance, the work of \cite{Gemp19Adam} considered optimising a cycle problem using Adam, which resembles the training procedure of GANs \cite{Goodfellow14GAN}. In \cite{ Zhang20AdamBetter}, the authors studied why Adam works better than the HB method when training Transformers by examining the gradient distributions.  

Insightful theoretical attempts have also been made to study the convergence behaviours of Adam. The incorrectness of the convergence proof of Adam in the original paper \cite{Kingma17} was spotted in  \cite{Bock18Adam, Rubio17Adam}. It was found in \cite{Reddi18Amsgrad} that Adam does not converge for a class of stochastic convex optimisation problems.  The recent work \cite{Bock19Adam} provides local convergence analysis for the update expressions (\ref{equ:Adam1})-(\ref{equ:Adam3_1}) of Adam from the viewpoint of discrete dynamic systems. The update expression (\ref{equ:Adam3_2}) does not fit into the analysis framework of   \cite{Bock19Adam}  due to the fact that $\sqrt{v}+\epsilon$ is not differentiable at $v=0$. Local convergence condition in terms of the parameters $(\eta, \epsilon, \beta_1)$ of Adam and the eigenvalues of Hessian of $f(x)$ is established. In brief, it is required that the learning rate $\eta$ needs to be sufficiently small 
to ensure a local optimal solution of $f(x)$ is asymptotically stable in a neighbourhood. The analysis results also hold for AdamW as will be discussed later on.

In this work, we propose Aida\footnote{it is named after an Italian opera by Verdi.}, a new optimisation method, by leveraging the 2nd moment and magnitude of AdamW. As is summarised in Alg.~1, the update direction $\frac{|m_{t+1}|^q\odot \textrm{sign}(m_{t+1})}{(r_{t+1} +\epsilon)^{q/p} }$ $\left(\textrm{or } \frac{|m_{t+1}|^q\odot \textrm{sign}(m_{t+1})}{r_{t+1}^{q/p} +\epsilon }\right)$ of Aida for computing $x_{t+1}$ is obtained by multiplying the magnitude $\frac{|m_{t+1}|^q}{(|r_{t+1}|+\epsilon)^{q/p}}$ $\left(\textrm{or } \frac{|m_{t+1}|^q}{r_{t+1}^{q/p} +\epsilon }\right)$ to the sign vector $\textrm{sign}(m_{t+1})$, where $r_{t+1}$ tracks the 2nd moment of the $p$th power of the gradient-magnitudes. When $p=2$ and $q=1$, the update equations reduce to those of AdamW as in (\ref{equ:Adam3_1})-(\ref{equ:Adam3_2}).  


The main objective for Aida was to obtain better local convergence, to be able to lessen the dependency on learning rate.
Suppose the origin 0 is a local optimal solution of a twice-differentiable function. It is found that when $q>1$ and $p>1$ in Aida, the origin 0 is asymptotically stable in a neighbourhood only when the weight decay $\mu$ is nonzero. There is no requirement on the learning rate $\eta$, which is verified by empirical investigation of a neighourhood of the optimal solution $x_{\ast}=0$ of $f(x)=\frac{10}{2}x^2$. Furthermore, the introduction of parameter $q$ makes it feasible to analyse the update expression involving $r^{q/p}+\epsilon$ when $q/p\geq 1$ (see Subsection~\ref{subsec:auto_H_b}).

The relaxation from $p=2$ of AdamW to $p\geq 1$ in Aida is to bring more flexibility of learning-rate adaptation to the optimisation method. As will be studied later on, the change from $p=2$ to $p=1$ makes the individual learning rates more aggressive, which is found to be beneficial in tested optimisation problems in the experimental section. Simulation results on ten toy optimisation problems\footnote{In the ten problems, the initial values $\{x_0\}$ are not close to their respective optimal solutions. This experiment was to evaluate the global convergence of Aida. } demonstrate that Aida with the setups $(p,q,)=(2,2)$ and $(p,q,)=(1,2)$ exhibit better convergence behaviours than $(p,q)=(2,1)$ in most tested problems.   


To study its effectiveness in practical applications,  Aida is evaluated on two deep learning tasks that are based on Transformers. The two tasks are: (1) training Transformers over the database WMT16:multimodal translation and (2) training Swin-Transformer over the database CIFAR10. It is found that in both tasks, Aida with the setup $(p,q)=(1,2)$ outperforms the setup $(p,q)=(2,1)$ (or equivalently AdamW) with a significant gain.  The above results indicate that Aida is not a trivial extension of AdamW or Adam and the exponent parameter $q$ affects the validation performance considerably for at least certain deep learning tasks.

\section{Related Work}


In the literature,  research on employing the sign vector $\textrm{sign}(g_t)$ of gradient in optimisation has received some attention.  In \cite{Riedmiller93signOpt}, the authors propose to adjust the update direction based on the change of gradient signs per iteration. The work \cite{Seide14signOpt} considers using stochastic gradient signs when training DNN models in a distributed setting to reduce communication bandwidth.  In \cite{Karimi16PLcon}, the authors provide new convergence results for using  $\textrm{sign}(g_t)$ when minimising a deterministic function. 

The recent work \cite{Balles18Adam} investigates Adam by reformulating the update expression (\ref{equ:Adam3_1}) or (\ref{equ:Adam3_2}) with $\mu=0$ in the form of sign and magnitude by ignoring the parameter $\epsilon$: 
\begin{align}
x_{t+1} &= x_{t} - \eta \frac{m_{t+1}}{\sqrt{v_{t+1}}} \nonumber \\
&= x_{t} - \eta \sqrt{\frac{1}{1 + \frac{v_{t+1}-m_{t+1}^2}{m_{t+1}^2} }} \odot \textrm{sign}(m_{t+1}). \nonumber
\end{align}
The ratio $\frac{v_{t+1}-m_{t+1}^2}{m_{t+1}^2}$ is interpreted as the estimated relative variance of gradients. The above reformulation is unrelated to our notion of magnitude modification of Adam and AdamW as shown in Alg.~1 for Aida.

\vspace{0mm}
\section{Preliminaries}
\vspace{-0mm}
\label{sec:problem}

In this section, we first present the optimisation problem considered in this paper.  We then briefly introduce the framework of discrete dynamic systems, which will be used later on to analyse the local convergence of Aida.  
\subsection{Problem definition}

In this work, we consider minimising a twice-differentiable function:
\begin{align}
 \min_{{x}\in \mathbb{R}^d} f({x}), \quad f \in C^2(\mathbb{R}^n, \mathbb{R}).
\label{equ:problem}
\end{align}
Suppose $x_{\ast}$ is a local optimal solution of (\ref{equ:problem}). Then we have 
\begin{align}
g(x_{\ast}) = 0 \quad \nabla_x g(x_{\ast})  \succ 0 \label{equ:opt},  
\end{align}
where $g(x) =\nabla_x f(x) $. The symbol $\succ$ in (\ref{equ:opt}) indicates that the Hessian matrix $\nabla_x g(x_{\ast})$ is symmetric positive definite. 


\subsection{Notion of discrete dynamical systems}
Inspired by \cite{Bock19Adam}, we will analyse the convergence of Aida later on by employing the framework of discrete dynamical systems.  By following the notation in \cite{Bock19Adam, Bof18dynamic}, we let $H: \mathbb{N}\times M\rightarrow \mathbb{R}^m$, $\mathbb{N}=\{0,1,\ldots\}$, $M\subset \mathbb{R}^m$, be a non-autonomous dynamical system in the form of
\begin{align}
z_{t+1} = H(t, z_t)\quad t \in \mathbb{N}. \label{equ:non-auto}
\end{align}
Denote the initial value $ z_0 \in M$ at iteration $t=0$. The notations $z_{t+1}=z_H(t;z_0)$ and $z=z_H(\cdot; z_0)$ indicate the dependence of the system states on $z_0$. 

An autonomous system refers to the scenario that $H$ does not depend on the iteration index $t$, which, for clarity purposes, is denoted as $\bar{H}: M\rightarrow   \mathbb{R}^m$ with the update form 
\begin{align}
z_{t+1} = \bar{H}( z_t)\quad t \in \mathbb{N}. \label{equ:auto}
\end{align}

Next we present the concept of local convergence for $\bar{H}$ via the introduction of fixed points,  local Lipschitz and asymptotical stability (see \cite{Bof18dynamic} for details).
\begin{definition}
$z_{\ast}\in M$ is called a fixed point of $H$ (or $\bar{H}$) if $H(t, z_{\ast})=z_{\ast}$ (or $\bar{H}( z_{\ast})=z_{\ast}$) for all $t\in \mathbb{N}$.  
\label{def:fixedpoint}
\end{definition}

\begin{definition}
A fixed point $z_{\ast}$ of $\bar{H}$ is asymptotically stable if there exists $\delta>0$ such that
\begin{align}
\|z_0-z_{\ast}\|_2 < \delta \Rightarrow \lim_{t\rightarrow \infty} z_{\bar{H}}(t;z_0) = z_{\ast}.
\end{align}
\label{def:asy_stable}
\end{definition}

\begin{definition}
The  system $\bar{H}$ in (\ref{equ:auto}) is locally Lipschitz on a domain $M\subset \mathbb{R}^m$ if each point $z_0$ in $M$ has a neighbourhood $M_0=\{z | \|z-z_0\|_2 \leq R \}$, $R>0$, such that
\begin{align}
\hspace{-3mm}\|\bar{H}(z_1) \hspace{-0.8mm}- \hspace{-0.8mm} \bar{H}(z_2) \|_2 \leq L_{z_0} \| z_1  \hspace{-0.8mm}- \hspace{-0.8mm} z_2 \|_2  \quad \forall z_1, z_2\in M_0,\label{equ:local_lip}
\end{align}
where $L_{z_0}>0$.
\label{def:local_lip}
\end{definition}

\begin{theorem}[Theorem~3.3 of \cite{Bof18dynamic}]
Let $z_{\ast}=0 \in M$ be a fixed point for the autonomous system (\ref{equ:auto}), where $\bar{H}:M\rightarrow \mathbb{R}^m$ is locally Lipschitz in $M\subset \mathbb{R}^m$. Let $J_{\bar{H}}(0)$ be the Jacobian matrix of $\bar{H}$ at $0$. Then $0$ is asymptotically stable if all the eigenvalues $\lambda_i|_{i=1}^m$ of $J_{\bar{H}}(0)$ satisfy $|\lambda_i| <1$, $i=1,\ldots, m$.  Instead, if there exists at least an eigenvalue such that $|\lambda_i | > 0$,
then the origin $0$ is unstable.
\label{theorem:auto_stable}
\end{theorem}

In a section later on, we will analyse the local convergence of Aida at a fixed point $z_{\ast}=0$ based on Theorem~\ref{theorem:auto_stable}. The key point is to investigate the eigenvalues of the Jacobian matrix $\bar{H}(0)$.


\section{Aida: A New Adaptive Optimisation Method Extending AdamW}
In this section, we first present the update expressions of Aida with two additional hyper-parameters $(p\geq 1,q\geq 1)$ in comparison to AdamW. It is noted that Aida reduces to AdamW when $(p,q)=(2,1)$. We then motivate the design of $q\geq 1$ in Aida by inspection of a convergence condition for AdamW, which is obtained by extending the convergence results in \cite{Bock19Adam} for analysing Adam. 

\subsection{Design of Aida adaptive method}
\noindent\textbf{Update expressions:} We attempt to extend the update expressions (\ref{equ:Adam1})-(\ref{equ:Adam3_2}) of AdamW in two aspects. Firstly, we propose to track the 2nd moment $r_t$ of the $p$th power of gradient magnitudes, which reduces to $v_{t}$ of the squared-gradient in AdamW when $p=2$. Secondly, we replace the magnitude vector $\frac{|m_{t+1}|}{\sqrt{v_{t+1}+\epsilon}}$ and $\frac{|m_{t+1}|}{\sqrt{v_{t+1}}+\epsilon}$ in (\ref{equ:Adam3_1})-(\ref{equ:Adam3_2}) by $\frac{|m_{t+1}|^q}{(r_{t+1}+\epsilon)^{q/p} }$ and $\frac{|m_{t+1}|^q}{r_{t+1}^{q/p}+\epsilon}$ before multiplying the sign vector $\textrm{sign}(m_{t+1})$. Hence, the update expressions of Aida can be represented as
{\small\begin{align}
\hspace{0mm} {m}_{t+1} & =\beta_1{m}_{t} + (1-\beta_1) g(x_t) \qquad \qquad  \qquad \qquad \label{equ:Aida1}\\
\hspace{0mm} {r}_{t+1} & = \beta_2 {r}_{t} + (1-\beta_2) |g(x_t)|^p  \qquad  \qquad  \qquad\qquad \label{equ:Aida2}
\end{align}}
\vspace{-6mm}
{\small \begin{subnumcases}{\hspace{-6mm}\textrm{or}}
\hspace{-2mm}{x}_{t+1} \hspace{-0.8mm}  =\hspace{-0.8mm}  (1\hspace{-0.8mm} -\hspace{-0.8mm} \mu) {x}_{t} \hspace{-0.6mm} - \hspace{-0.6mm}   \eta \frac{ (1  \hspace{-0.6mm}  -\hspace{-0.6mm}   \beta_2^{t+1})^{q/p} }{(1 \hspace{-0.6mm}  -\hspace{-0.8mm}    \beta_1^{t+1})^q }\hspace{-0.6mm}   \frac{|m_{t\hspace{-0.3mm}+\hspace{-0.3mm}1}|^q\textrm{sign}(m_{t\hspace{-0.3mm}+\hspace{-0.3mm}1}) }{(r_{t+1}+\epsilon)^{q/p} } \label{equ:Aida3_1} \\
\hspace{-2mm}{x}_{t+1} \hspace{-0.8mm}  =\hspace{-0.8mm}   (1\hspace{-0.8mm} -\hspace{-0.8mm} \mu){x}_{t} \hspace{-0.6mm} - \hspace{-0.6mm}   \eta \frac{ (1   \hspace{-0.8mm}  -\hspace{-0.8mm}     \beta_2^{t+1})^{q/p} }{(1 \hspace{-0.6mm}  -\hspace{-0.8mm}   \beta_1^{t+1})^q }\hspace{-0.6mm}   \frac{|m_{t\hspace{-0.3mm}  +\hspace{-0.3mm}  1}|^q\textrm{sign}(m_{t\hspace{-0.3mm}  +\hspace{-0.3mm}  1}) }{ r_{t+1}^{q/p}+\epsilon }, \label{equ:Aida3_2} 
\end{subnumcases}}
\hspace{-1.5mm}where $g(x_t)=\nabla f({x}_{t})$, $p\geq 1$, $q\geq1$. It is immediate that when $(p,q)=(2,1)$, (\ref{equ:Aida1})-(\ref{equ:Aida3_2}) reduce to (\ref{equ:Adam1})-(\ref{equ:Adam3_2}) of AdamW.  See Alg.~\ref{alg:Aida} for a summary of Aida for optimisation of a deterministic function $f(x)$. 

We now consider the impact of $q$ by fixing $p=2$ in (\ref{equ:Aida1})-(\ref{equ:Aida3_1}) as an example. When $q=1$, the magnitude vector $\frac{|m_{t+1}|}{\sqrt{r_{t+1}+\epsilon}}$ indicates that those coordinates of $\textrm{sign}(m_{t+1})$ with magnitudes $|m_{t+1}|$ larger than the corresponding elements of $\sqrt{r_{t+1}+ \epsilon}$ would receive learning rates that are greater than 1, and tend to be aggressive when updating their corresponding coordinates of ${x}$. On the other hand, the remaining coordinates of  $\textrm{sign}(m_{t+1})$ would receive learning rates that are smaller than or equal to 1 and tend to be conservative. When $q>1$, the magnitude vector becomes $\frac{|m_{t+1}|^q}{(\sqrt{r_{t+1}+\epsilon})^q}$. The change from $q=1$ to $q>1$ increases those learning rates that are greater than 1 and decreases the remaining ones. In the next subsection, we will motivate the choice of $q>1$ in Aida from a local convergence point of view. 

Next we study the impact of the change from $p=2$ to $p=1$ on the 2nd moment $\{r_t| t\geq 0\}$. By employing Jensen's inequality, it is immediate that  
\begin{align}
\sqrt{\frac{r_{t,[i], p=2}}{(1-\beta_2^{t}) } } \geq  \frac{r_{t,[i], p=1}}{(1-\beta_2^{t}) },  \label{equ:ineq_p}
\end{align} 
where the subscript $[i]$ indicates the $i$th component of $r_t$, $\frac{1}{1-\beta_2^t}$ is the normalising factor. (\ref{equ:ineq_p}) indicates that when $p=1$ is selected instead of $p=2$, an increasing number of coordinates of $\textrm{sign}(m_{t+1})$ would receive learning rates that are greater than 1, making the updates of Aida more aggressive per-iteration. In the experiment later on, we mainly study the setups of $p=\{1,2\}$ and $q=\{1,2\}$.  


\noindent\textbf{Reformulation as discrete dynamic systems:} Suppose  $\{x_t\}$, $\{m_t\}$, and $\{r_t\}$ are computed by following either $\{(\ref{equ:Aida1}), (\ref{equ:Aida2}), (\ref{equ:Aida3_1})\}$ or $\{(\ref{equ:Aida1}), (\ref{equ:Aida2}), (\ref{equ:Aida3_2})\}$.  We let $z_{t} = [m_{t}^T, r_{t,}^T, x_{t}^T]^T\in \mathbb{R}^{3n} $ for all $t\geq 0$.  

We first consider the update expressions $\{(\ref{equ:Aida1}), (\ref{equ:Aida2}), (\ref{equ:Aida3_1})\}$. By following a similar analysis as that used in \cite{Bock19Adam} for Adam, the  expressions can be rewritten as the iteration of a combination of a non-autonomous and an autonomous dynamic system as 
\begin{align}
z_{t+1} =H_a(t,z_t)= \bar{H}_a(z_t) +\Theta_a(t,z_t), \label{equ:dynAida}
\end{align} 
where the subscript $a$ indicates the use of the update form (\ref{equ:Aida3_1}) for $x$, and
\begin{align}
\bar{H}_a(z_t) = \left[\begin{array}{c} 
\beta_1 m_t +(1-\beta_1)g(x_t) \\ 
\beta_2 r_t +(1-\beta_2)|g(x_t)|^p \\  
(1-\mu)x_t-\eta \frac{|m_{t+1}|^q \odot \textrm{sign}(m_{t+1})}{ ({r_{t+1}+\epsilon})^{q/p}}
  \end{array}\right], \label{equ:autoAida}
\end{align} 
and 
\begin{align} 
\hspace{-2.5mm}{\Theta}_a(t, z_t) \hspace{-1mm} =\hspace{-1.5mm} \left[\hspace{-1.5mm}  \begin{array}{c} 0 \\ 0 \\ -\eta\left(\frac{ (1-\beta_2^{t+1})^{p/q} }{(1-\beta_1^{t+1})^q } \hspace{-0.6mm} -\hspace{-0.6mm}  1\right) \hspace{-1.2mm}  \frac{|m_{t+1}|^q\odot \textrm{sign}(m_{t+1})}{ (r_{t+1}+\epsilon)^{q/p} } \end{array} \hspace{-1.5mm}  \right] \hspace{-1.2mm}. \hspace{-2mm}\label{equ:nonautoAida}
\end{align}

Similarly, we denote the dynamic system for the update expressions $\{(\ref{equ:Aida1}), (\ref{equ:Aida2}), (\ref{equ:Aida3_2})\}$ as 
\begin{align}
z_{t+1} =H_b(t,z_t)= \bar{H}_b(z_t) +\Theta_b(t,z_t), \label{equ:dynAida_2nd}
\end{align} 
where the expressions for $\bar{H}_b(z_t)$ and $\Theta_b(t,z_t)$ can be easily derived, which we omit here to save space.

The lemma below characterises the relationship between a local optimal solution $x_{\ast}$ of $f$ and a fixed point of $H_a(t,z)$ or $H_b(t,z)$:
\begin{lemma}
Suppose  $x_{\ast}$  is a local optimal solution of $f$ satisfying (\ref{equ:opt}). Then $z_{\ast}=(0^T,0^T,x_{\ast}^T)^T$ is a fixed point of $H_a(t,x)$ in (\ref{equ:dynAida}) or  $H_b(t,x)$ in (\ref{equ:dynAida_2nd}). That is, $\bar{H}_a(z_{\ast})=z_{\ast}$, $\bar{H}_b(z_{\ast})=z_{\ast}$,  $\Theta_a(t,z_{\ast})=z_{\ast}$,  $
\Theta_b(t,z_{\ast})=z_{\ast}$ for all $t\geq 0$.
\label{lemma:fixpoint}
\end{lemma} 

Similarly to \cite{Bock19Adam}, we will attempt to identify conditions  under which a fixed point $z_{\ast}$ in Lemma~\ref{lemma:fixpoint} is asymptotically stable for the autonomous system $\bar{H}_a$ (or $\bar{H}_b$)  of Aida. 




\subsection{Motivation of $q\geq 1$ in Aida over $q=1$ in AdamW via convergence analysis of $\bar{H}_a(z_t)$}

In this subsection, we will analyse the local convergence of $\bar{H}_a(z_t)$ in (\ref{equ:dynAida}) in a neighbourhood of a fixed point $z_{\ast}=[m_{\ast}^T,r_{\ast}^T, x_{\ast}^T]^T=0$. The investigation for $\bar{H}_b(z_t)$ will be postponed to next subsection due to fact that \cite{Bock19Adam} only focuses on $\bar{H}_a(z_t)$ with $(p,q)=(2,1)$ and $\mu=0$. 

By inspection of Theorem~\ref{theorem:auto_stable}, it is clear that the key step is to identify conditions under which the spectral radius  of the Jacobian $J_{\bar{H}_a}(z)$ of $\bar{H}_a(z)$ at $z=0$ is less than 1 (i.e., $\rho(J_{\bar{H}_a}(0)<1$). In the following, the expression for  $J_{\bar{H}_a}(z_t)$ is derived first and the  continuity of its elements over $z_t$ will be studied. 
After that, we consider $(p,q)=(2,1)$ for $\bar{H}_a(0)$ and derive a condition that leads to $\rho(J_{\bar{H}_a}(0)<1$ by extending the results in \cite{Bock19Adam} from Adam ($\mu=0$ in (\ref{equ:Adam1})-(\ref{equ:Adam3_2}) ) to AdamW (i.e., $\mu>0$ in (\ref{equ:Adam1})-(\ref{equ:Adam3_2})). We then show that when $q$ changes from $q=1$ to $q>1$, a less strict condition on the learning rate $\eta$ can be derived for the general scenario $p> 1$ which includes $p=2$ as a special case.  



\noindent\textbf{Jacobian of $\bar{H}_a(z_t)$}: By inspection of (\ref{equ:autoAida}), the Jacobian matrix $J_{\bar{H}_a}(z_t)$ of $\bar{H}_a(z_t)$ takes the form of 
{\small 
\begin{align}
\hspace{-5mm}J_{\bar{H}_a}(z_t) \hspace{-1mm}=\hspace{-1mm} \left(\begin{array} {ccc} 
\beta_1I &  0 & (1\hspace{-0.8mm} -\hspace{-0.8mm} \beta_1) \nabla_x g(x_t) \\
0 &   \beta_2 I  & \frac{\partial r_{t+1}}{ \partial x_t }  \\
\frac{\partial x_{t+1}}{ \partial m_t } & \frac{\partial x_{t+1}}{ \partial r_t } & \frac{\partial x_{t+1}}{ \partial x_t }
 \end{array}\right),  \hspace{-3mm} \label{equ:J_t_Aida}
\end{align}
}
where  
\begin{align}
 \frac{\partial r_{t+1}}{ \partial x_t }  &=  p(1-\beta_2)\textrm{diag}(|g(x_t)|^{p-1}\odot\textrm{sign}(g_t)) \nabla_x g(x_t)   \nonumber \\
 \frac{\partial x_{t+1}}{ \partial m_t }  &=  -\eta \textrm{diag}\left(\frac{q\beta_1| m_{t+1}|^{q-1}}{ (r_{t+1}+\epsilon)^{q/p} } \right)  \nonumber \\
  \frac{\partial x_{t+1}}{ \partial r_t }  &= \eta\textrm{diag}\left( \frac{(p/q)  \beta_2 | m_{t+1}|^q \textrm{sign}(m_{t+1}) }{ ( r_{t+1}+\epsilon)^{q/p+1} } \right) \nonumber \\
   \frac{\partial x_{t+1}}{ \partial x_t }  &= (1-\mu) I - \eta \textrm{diag}\Big[\frac{ q|m_{t+1}|^{q-1}(1-\beta_1) }{ (r_{t+1}+\epsilon)^{q/p} }  \nonumber \\
  &\hspace{-8mm} -\frac{ q|m_{t+1}|^{q}\textrm{sign}(m_{t+1}) (1\hspace{-0.4mm}-\hspace{-0.4mm}\beta_2)|g(x_t)|^{p-1}\textrm{sign}(g(x_t)) }{(r_{t+1}+\epsilon)^{q/p+1} }  \Big] \nonumber \\
  & \cdot  \nabla_x g(x_t), \nonumber
\end{align}
where it is noted again that $\nabla_x g(x_t)$ is the Hessian matrix of $f$ at $x_t$. 

Note that the convergence results of Theorem~\ref{theorem:auto_stable} hold only when the considered system $\bar{H}$ is locally Lipschitz. The lemma below describes the relationship between the continuity of the Jacobian $J_{\bar{H}}$ of $\bar{H}$ and the system  $\bar{H}$ being locally Lipschitz.
 
\begin{lemma}
If an autonomous system $\bar{H}:M\rightarrow \mathbb{R}^m$, $M\in\mathbb{R}^{m}$, is continuously differentiable in a domain $M$ (or equivalently, its Jacobian matrix $J_{\bar{H}}$ is continuous in $M$ ), then it is locally Lipschitz.  
\label{lemma:conti_Lips}
\end{lemma}

We now consider the functional continuity of $J_{\bar{H}_a}(z_t)$ in (\ref{equ:J_t_Aida}).  It is known that the sign operator $\textrm{sign}(y)$, $y\in \mathbb{R}$, is discontinuous at $y=0$.  The quantity $\textrm{sign}(0)$ may take any value in the range $[-1,1]$. Fortunately, if $(p, q)$ are configured to satisfy $p>1$ and $q\geq 1$,  it is not difficult to show that $\bar{H}_a(z_t)$ is continuous in $\mathbb{R}^{3n}$, thus satisfying the condition of Lemma~\ref{lemma:conti_Lips}. The remaining work is to study the spectral radius of $J_{\bar{H}_a}(z)$ at a fixed point $z=0$. 


\noindent\textbf{Convergence condition for $\bar{H}_a(z_t)$ with $(p,q)=(2,1)$}:
Suppose $z=0$ is a fixed point of $\bar{H}_a(z_t)$ when $(p,q)=(2,1)$. By setting $(p,q)=(2,1)$ in (\ref{equ:J_t_Aida}) and using the property that  $ (m_{\ast}, r_{\ast},g(0))=(0,0,0)$, the Jacobian matrix $J_{\bar{H}_a}(0)$ can be represented as
{\small\begin{align}
 &J_{\bar{H}_a}(0)  \hspace{-1.2mm} \nonumber \\
 &= \hspace{-1.2mm} \left(\begin{array}{ccc} \beta_1 I & 0 &  (1-\beta_1 )\nabla_x g(0) \\
0 & \beta_2 I & 0 \\
 -\frac{\eta\beta_1}{\sqrt{\epsilon}} I & 0 &  (1-\mu)I - \frac{\eta (1-\beta_1)}{\sqrt{\epsilon}}\nabla_x g(0) \\
\end{array} \right)  \hspace{-1.2mm}. \hspace{-1.5mm} \label{equ:J_opt_Adam}
\end{align}}  

The work \cite{Bock19Adam} investigates the spectral radius of $J_{\bar{H}_a}(0)$ in (\ref{equ:J_opt_Adam}) for the special case $\mu=0$ (which is Adam). It is straightforward to extend the results of \cite{Bock19Adam} for the general case $\mu\geq 0$ in  (\ref{equ:J_opt_Adam}), which are summarised in a theorem below: 
\begin{theorem}[Variant of Theorem III.2 and 3 in \cite{Bock19Adam}]
Consider $\bar{H}(z_t)$ of (\ref{equ:J_t_Aida}) with $p=2$ and $q=1$. Let $x_{\ast}=0\in \mathbb{R}^n$ be a minimum of $f$ with a positive definite Hessian $\nabla_x^2 f(0)=\nabla_x g(0)$. Denote $\gamma_i|_{i=1}^n$ and $\lambda_i|_{i=1}^{3n}$ to be the eigenvalues of $\nabla_x g(0)$ and $J_{\bar{H}}(0)$, respectively. Then $\lambda_i|_{i=1}^{3n}$ can be represented as
{\small \begin{align}
\lambda_{1,i} =& \beta_2 \label{equ:lambda1_J_Adam}\\
\lambda_{2,i} =&  \frac{1 \hspace{-0.7mm}- \mu \hspace{-0.7mm}+\hspace{-0.7mm}\beta_1 \hspace{-0.7mm}-\hspace{-0.7mm} \varphi_i(1-\beta_1)}{2} \nonumber \\
& +  \frac{\hspace{-0.7mm}\sqrt{(1\hspace{-0.7mm}-\hspace{-0.7mm}\mu\hspace{-0.7mm}+\hspace{-0.7mm}\beta_1\hspace{-0.7mm}-\hspace{-0.7mm}\varphi_i(1-\beta_1) )^2\hspace{-0.7mm}-\hspace{-0.7mm}4\beta_1(1\hspace{-0.7mm}-\hspace{-0.7mm}\mu) }}{2} \label{equ:lambda2_J_Adam} \\
\lambda_{2,i} =&  \frac{1 \hspace{-0.7mm}- \mu \hspace{-0.7mm}+\hspace{-0.7mm}\beta_1 \hspace{-0.7mm}-\hspace{-0.7mm} \varphi_i(1-\beta_1)}{2} \nonumber \\
& -  \frac{\hspace{-0.7mm}\sqrt{(1\hspace{-0.7mm}-\hspace{-0.7mm}\mu\hspace{-0.7mm}+\hspace{-0.7mm}\beta_1\hspace{-0.7mm}-\hspace{-0.7mm}\varphi_i(1-\beta_1) )^2\hspace{-0.7mm}-\hspace{-0.7mm}4\beta_1(1\hspace{-0.7mm}-\hspace{-0.7mm}\mu) }}{2}  \label{equ:lambda3_J_Adam} 
\end{align} }
\hspace{-2mm}where $\varphi_i = \frac{\eta\gamma_i}{\sqrt{\epsilon}}$. The spectral radius $\rho(J_{\bar{H}}(0))<1$ if the following condition holds  
\begin{align}
\frac{\eta \max_{i=1}^n \gamma_i }{\sqrt{\epsilon}}(1-\beta_1) < 2\beta_1+2(1-\mu). \label{equ:cond_Adam}
\end{align}
\label{theorem:cond_Adam}
\end{theorem}

The condition (\ref{equ:cond_Adam}) involves the parameters $\eta$, $\epsilon$, $\beta_1$, $\mu$, and the eigenvalues $\gamma_i|_{i=1}^n$ of the Hessian matrix $\nabla_x^2 f(0)$. The inequality indicates that for a setup $\beta_1=0.9$ and the maximum eigenvalue $\max_{i=1}^n \gamma_i$ being of order 1(i.e., $ \mathcal{O}(\max_{i=1}^n \gamma_i)=1$),  the learning rate $\eta$ has to be  either smaller or comparable to $\sqrt{\epsilon}$ to ensure local stability at the optimal solution $z_{\ast}=0$.  

\begin{figure}[t!]
\centering
\includegraphics[width=60mm]{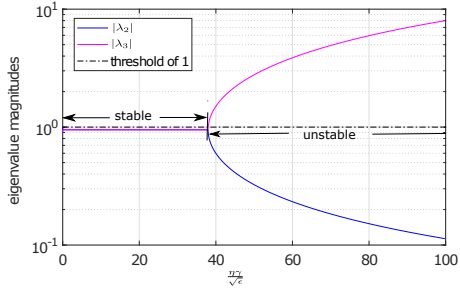}
\caption{\small Visualisation of the magnitudes of the two eigenvalues $\lambda_2$ and $\lambda_3$ of (\ref{equ:lambda2_J_Adam})-(\ref{equ:lambda3_J_Adam})  as a function of $\frac{\eta \gamma}{\sqrt{\epsilon}}$ by setting $\beta_1=0.9$ and $\mu=1e-5$, where the subscript $i$ is dropped for simplicity. } 
\label{fig:Aida_eigen}
\end{figure}

Fig.~\ref{fig:Aida_eigen} visualises the eigenvalue magnitudes of $\lambda_{2}$ (subscript $i$ is dropped) and $\lambda_{3}$ as a function of $\varphi$ with $(\beta_1,\mu)=(0.9, 1e-5)$, which demonstrates that the maximum value of $\eta\gamma$ has to be comparable to $\sqrt{\epsilon}$ to make $\rho(J_{\bar{H}}(0))<1$. We will show later that the change from $q=1$ to $q>1$ would lead to a much simpler convergence condition than (\ref{equ:cond_Adam}).

\noindent\textbf{Simpler convergence condition for $\bar{H}(z_t)$ with $q>1$}:  Suppose $z=0$ is a fixed point of $\bar{H}_a(z_t)$. The Jacobian matrix $J_{\bar{H}_a}(z_t)$ of (\ref{equ:J_t_Aida}) at $z_t=0$ takes a simple form with $q>1$
{\small\begin{align}
 J_{\bar{H}_a}(0) 
 &= \hspace{-1.2mm} \left(\begin{array}{ccc} \beta_1 I & 0 &  (1-\beta_1 )\nabla_x g(0) \\
0 & \beta_2 I & 0 \\
0 & 0 &  (1-\mu) I
\end{array} \right). \hspace{-1.2mm} \label{equ:J_opt_Aida}
\end{align}}  
\hspace{-1.2mm}In comparison to (\ref{equ:J_opt_Adam}), the setup $q>1$ results in a few additional elements in (\ref{equ:J_opt_Aida}) being zero due to the quantity $|m_{t+1}|^q$ in (\ref{equ:Aida3_1}). 

Note that $ J_{\bar{H}_a}(0) $ in (\ref{equ:J_opt_Aida}) is an upper triangular matrix. Its diagonal elements are also the eigenvalues of the matrix. Its corresponding local convergence condition can be easily derived in terms of $\beta_1$, $\beta_2$ and $\mu$: 
\begin{theorem}
Consider $\bar{H}(z_t)$ of (\ref{equ:J_t_Aida}) with $p\geq 1$ and $q>1$. Let $x_{\ast}=0\in \mathbb{R}^n$ be a minimum of $f$. Then the eigenvalues $\lambda_i|_{i=1}^{3n}$ of  $J_{\bar{H}_a}(0)$ are  
\begin{align}
\lambda_{1,i}=\beta_{1}, \quad \lambda_{2,i}=\beta_{2}, \quad  \lambda_{3,i}=1-\mu. \label{equ:eigen_Aida}
\end{align}
The spectral radius  $\rho(J_{\bar{H}}(0))<1$ if the weight-decay parameter $\mu$ satisfies
\begin{align}
1>\mu>0.  \label{equ:cond_Aida}
\end{align}
\label{theorem:cond_Aida}
\vspace{-5mm}
\end{theorem}

It is clear that the condition (\ref{equ:cond_Aida}) for $q>1$ is much simpler than (\ref{equ:cond_Adam}) for $(p,q)=(2,1)$.  There is no explicit restriction on the learning rate $\eta$ in (\ref{equ:cond_Aida}) in terms of $\epsilon$, $\beta_1$ and the eigenvalues of the Hessian matrix $\nabla_x^2f(0)$. This suggests that when $q>1$, a relative large learning rate $\eta$ might be acceptable to ensure local stability.  

It is also worth noting that the weight-decay $\mu$ can be set to $\mu=0$ in the condition (\ref{equ:cond_Adam}) without affecting the overall local convergence property. On the other hand, the condition (\ref{equ:cond_Aida}) for $q>1$ requires $\mu>0$ to ensure local convergence. To briefly summarise, the change from $q=1$ to $q>1$ results in the transfer of the restriction on $\eta$ to a restriction on $\mu$. 

\begin{remark}
\label{remark:mu_0}
For the case that $\mu=0$, the eigenvalues $\lambda_{3,i}|_{i=1}^n=1$ in (\ref{equ:eigen_Aida}). It is undefined from Theorem~\ref{theorem:auto_stable} if $z_{\ast}=0$ is stable or not in its neighbourhood. As will be discussed later on, our empirical results (See Table~\ref{tab:toy} and Fig.~\ref{fig:Aida_toy}) indicate that the setup $\mu=0$ and $(p,q)\neq (2,1)$ in Aida gives satisfactory convergence results in comparison to Adam for optimisation problems of which the optimal solutions $z_{\ast}\neq 0$ and $z_{\ast}= 0$.    
\end{remark} 

 
Finally, combining Lemma~\ref{lemma:conti_Lips},  Theorem~\ref{theorem:auto_stable} and \ref{theorem:cond_Aida}, we obtain the following local convergence results for $\bar{H}_{a}(z_t)$ at a fixed point $z_{\ast}=0$: 
 
\begin{theorem}
Consider $\bar{H}_a(z_t)$ of (\ref{equ:autoAida}) with $p> 1$ and $q>1$. Let $x_{\ast}=0\in \mathbb{R}^n$ be a minimum of $f$. If $\mu$ satisfies (\ref{equ:cond_Aida}),  then $0$ is asymptotically stable in a neighbourhood of $0$.   
\label{theorem:converge_Aida_a}
\vspace{-1mm}
\end{theorem}

\subsection{Local convergence for $\bar{H}_{b}(z_t)$}
\label{subsec:auto_H_b}
By inspection of (\ref{equ:Aida3_1}) and (\ref{equ:Aida3_2}), it is noted that $\bar{H}_{b}(z_t)$ employs the quantity $r^{q/p}+\epsilon$ while $\bar{H}_{a}(z_t)$ uses $(r+\epsilon)^{q/p}$. That is, the parameter $\epsilon$ is treated differently in the two dynamic systems. Considering $r^{q/p}+\epsilon$, it is clear that when $q/p\geq 1$, the quantity is differentiable at $r=0$.  Accordingly, all the analysis for $\bar{H}_{a}(z_{t})$ can be applied to $\bar{H}_{b}(z_t)$. We summarise the results in a theorem below: 

\begin{theorem}
Consider $\bar{H}_b(z_t)$ with $p> 1$ and $q>1$. Let $x_{\ast}=0\in \mathbb{R}^n$ be a minimum of $f$. If $\mu$ satisfies (\ref{equ:cond_Aida}),  then $0$ is asymptotically stable in a neighbourhood of $0$.   
\label{theorem:converge_Aida_b}
\vspace{-1mm}
\end{theorem}


\section{Experiments}
In this section, we first verify the obtained local convergence condition for Aida in a neighbourhood of $x_{\ast}=0$ of the quadratic function $f(x)=\frac{10}{2}x^2$. We then evaluate the global convergence (i.e., $x_0$ is far away from $x_{\ast}$) of Aida for solving ten toy optimisation problems from \cite{Andrei08optFamlily}. After that, we consider real applications of Aida by training Transformer and its variant.    

\subsection{On local convergence of Aida}
Suppose the objective function is $f(x)=\frac{10}{2}x^2$, $x\in \mathbb{R}$, with the optimal solution $x_{\ast}=0$. Its Hessian matrix is constant, i.e., $\nabla_x^2 f(x)=10$. Accordingly, the eigenvalue of the Hessian matrix is $\gamma = 10$. We let the initial value $x_{0}=1e-10$, which has a very small error distance with respect to $x_{\ast}$ as $\|x_0-x_{\ast}\|_2=1e-10$.

 \begin{figure}[t!]
\centering
\includegraphics[width=80mm]{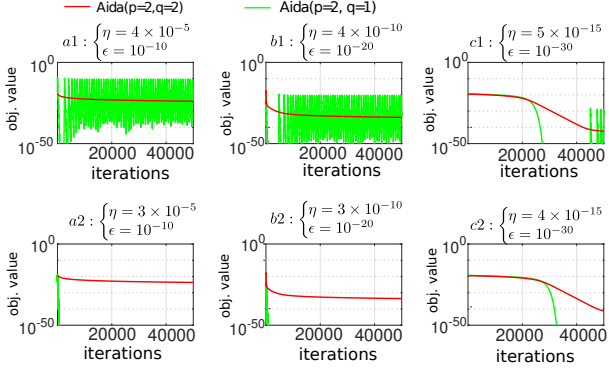}
\caption{\small Impact of $\eta$ and $\epsilon$ on local convergence of $\bar{H}_a(z_t)$ with $(p,q)=(2,2)$ and $(p,q)=(2,1)$  for minimising $f(x)=\frac{10}{2}x^2$, where $x_0$ is initialised to be $x_0=1e-10$.  The other parameters are set to be $(\beta_1, \beta_2, \mu)= (0.9, 0.999, 1e-10)$. Note that the Hessian of $f(x)$ is a scalar value 10, and therefore the eigenvalue $\gamma$ in Fig.~\ref{fig:Aida_eigen} is $\gamma=10$.   } 
\label{fig:Aida_quad}
\vspace{-2mm}
\end{figure}

\noindent\textbf{Impact of $\eta$ and $\epsilon$}: We first consider the impact of $\eta$ and $\epsilon$ on the local convergence of $\bar{H}_a(z_t)$ with $(p,q)=(2,1)$ and $(p,q)=(2, 2)$. The other parameters are set to be $(\beta_1, \beta_2, \mu)= (0.9, 0.999, 1e-10)$. The convergence results are displaced in Fig.~\ref{fig:Aida_quad}. This experiment is mainly for verifying the local convergence condition (\ref{equ:cond_Adam}) for the setup $(p,q)=(2,1)$.

The bottom three plots $\{a2, b2, c2\}$ are for the scenarios that with proper learning rate selection for different $\epsilon$, the setup $(p,q)=(2, 1)$ (which is AdamW) exhibits local stability around $x_{\ast}=0$. In the three plots, the product $\gamma \eta$ is roughly equal to $\sqrt{\epsilon}$ as $\epsilon$ decreases from $10^{-10}$ until $10^{-30}$, which is consistent with the analysis results in Theorem~\ref{theorem:cond_Adam}.  

The top three plots  $\{a1, b1, c1\}$ in Fig.~\ref{fig:Aida_quad} indicate that when the learning rate $\eta$ increases slightly with regard to those values in $\{a2, b2, c2\}$, the setup $(p,q)=(2, 1)$ exhibits unstable behaviours (i.e., fluctuates around certain value) in a neighbourhood of $x_{\ast}=0$. Additional experiments on minimising $f(x)=\frac{1}{2}x^2$ suggest that when $(p,q)=(2, 1)$, the proper learning rate to ensure local convergence also depends on the eigenvalue of the Hessian matrix $\nabla_x^2 f(x_{\ast})$.

Fig.~\ref{fig:Aida_quad} also shows that for all six plots, the setup $(p,q)=(2,2)$ exhibits stable behaviours (i.e., does not fluctuate around certain value) in a neighbourhood of $x_{\ast}=0$. Furthermore, as $\epsilon$ decreases, the setup $(p,q)=(2,2)$ converges to a point which is increasingly closer to the optimal solution $x_{\ast}=0$.

\begin{figure}[t!]
\centering
\includegraphics[width=80mm]{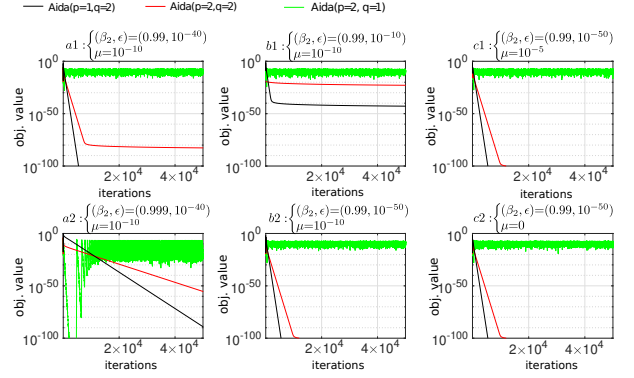}
\caption{\small  Impact of $\beta_2$, $\epsilon$, and $\mu$ on local convergence of $\bar{H}_a(z_t)$ for minimising $f(x)=\frac{10}{2}x^2$, where $x_0$ is initialised to be $x_0=1e-10$. The other parameters are set as $(\beta_1, \eta)= (0.9, 1e-3)$. } 
\label{fig:Aida_quad_beta_eta}
\vspace{-2mm}
\end{figure}


\noindent\textbf{Impact of $\beta_2$, $\epsilon$, and $\mu$}: Next we fix $(\beta_1, \eta)=(0.9, 1e-3)$ and investigate the impact of $\beta_2$, $\epsilon$, and $\mu$ on the local convergence of $\bar{H}_a(z_t)$. It is noted that the learning rate value $\eta=1e-3$ is a typical setup in the original paper of Adam \cite{Kingma17}.  This experiment is mainly for showing that when $q>1$, relatively large learning rate is indeed acceptable to enable local stability of $\bar{H}_a(z_t)$ around $x_{\ast}=0$.

Fig.~\ref{fig:Aida_quad_beta_eta} displays the local convergence results for different configurations of $(\beta_2, \epsilon, \mu)$. Let us first focus on the setup $(p,q)=(2,2)$. Interestingly, it is found from the two pairs of plots $a1-a2$ and $b1-b2$ that $\beta_2$ affects the convergence speed while $\epsilon$ affects the minimum level that Aida could reach. It is preferable to set a relatively small $\beta_2$ value (e.g., $\beta_2=0.99$) for $(p,q)=(2,2)$ compared to the recommended setup $\beta_2=0.999$ in \cite{Kingma17} for Adam (i.e., $(p,q)=(2,1)$).  In general, a small $\beta_2$ value allows the 2nd moment $\{r_t\}$ of Aida to easily adapt to the change of recent gradient-magnitudes over iterations.

The last pair of plots $c1-c2$ in Fig.~\ref{fig:Aida_quad_beta_eta} suggest that the parameter $\mu$ does not really affect the local convergence behaviour of $\bar{H}_a(z_t)$ with $q>1$ for the considered quadratic problem. In particular, the zero weight-decay $\mu=0$ does not lead to divergence of $\bar{H}_a(z_t)$ with $q>1$. As stated in Remark~\ref{remark:mu_0}, $\mu=0$ will make a number of eigenvalues of the Jacobian matrix $J_{\bar{H}_a(0)}$ being 1. In this case, its stability around $z_{\ast}=0$ remains undefined in Theorem~\ref{theorem:auto_stable}.

One can also conclude from the figure that the setup $(p,q)=(1,2)$ converges faster than $(p,q)=(2,2)$. This might be because when $p=1$, the updates of Aida tend to be more aggressive (see the inequality (\ref{equ:ineq_p}) which discusses the relationship between $\{r_{t,p=1}\}$ and $\{r_{t,p=2}\}$).

The analysis above regarding Fig.~\ref{fig:Aida_quad}-\ref{fig:Aida_quad_beta_eta} indicates that the change from $q=1$ to $q=2$ makes a large impact on the local convergence of Aida in a neighbourhood of $x_{\ast}$ for the considered quadratic optimisation problem. The algorithm with $q=2$ tends to be more stable than the setup $q=1$. This is in line with our stability analysis of $\bar{H}_a(z_t)$ in early sections.  

\subsection{On global convergence of Aida over toy problems}
In this subsection, we consider applying Aida for solving ten toy optimisation problems as listed in Table~\ref{tab:toy}, where both the functions and the initialisations for $x_0$ are from \cite{Andrei08optFamlily}. All the functions are twice-differentiable, and their Hessian matrices exist.  Based on results of the first experiment, the parameters of Aida except $(p,q)$ were set to $(\beta_1, \beta_2, \mu, \eta, \epsilon)= (0.9, 0.99, 0, 0.001, 1e-50)$.  

Note from Table~\ref{tab:toy} that the optimal solutions $x_{\ast}$ of the ten problems are not always the origin $0$. For example, $x_{\ast}$ of TRIDIA is $x_{\ast}=[1, \frac{1}{2}, \ldots, \frac{1}{2^{n-1}}]$. EDENSCH also has a nonzero optimal solution. The objective of this experiment is to investigate the global convergence (i.e., $x_0$ is far away from $x_{\ast}$) of Aida with $(p,q)\neq(2,1)$ for solving different types of problems, which is uncovered in our earlier local convergence analysis.

\begin{table}[t]
\caption{\small  List of 10 optimisation problems from \cite{Andrei08optFamlily}. The "[]" symbol in the subscript indicates a particular component of  a variable $x\in \mathbb{R}^n$.  } 
\label{tab:toy}
\centering
\begin{tabular}{|c|c|c|}
\hline
 {{\scriptsize func.  }} & \hspace{0mm} {{\scriptsize func. expression }}\hspace{0mm}
   & \hspace{0mm}{\scriptsize  $x_0$ }     \hspace{0mm} 
 \\
\hline
\hline
 \hspace{-6mm} \scriptsize{ $\begin{array}{c}\textrm{fun1:} \\ \textrm{Ext. Himmelblau:} \end{array}$} \hspace{-4.5mm} & \hspace{-4.5mm}  {\scriptsize $\begin{array}{l} \sum_{i=1}^{n/2} (x_{[2i-1]}^2 \hspace{-0.8mm} +\hspace{-0.8mm}   x_{[2i]}  \hspace{-0.8mm} -\hspace{-0.8mm}  11) \\ +\sum_{i=1}^{n/2} (x_{[2i-1]}  \hspace{-0.8mm} +\hspace{-0.8mm}  x_{[2i]}^2 \hspace{-0.8mm} -\hspace{-0.8mm}  7)^2 \end{array}$  }  \hspace{-4.5mm} &  \hspace{-4.5mm}  \scriptsize{ $\begin{array}{c} n=100 \\ \textrm{[1,1,$\ldots$, 1$]^T$}  \end{array}$ }   \hspace{-5mm}
\\ \hline
 \hspace{-6mm} \scriptsize{ $\begin{array}{c}\textrm{fun2:} \\ \textrm{Ext. Maratos:} \end{array}$} \hspace{-4.5mm} & \hspace{-4.5mm}  {\scriptsize $\begin{array}{l} \sum_{i=1}^{n/2}x_{[2i-1]} +100 \\ \cdot \sum_{i=1}^{n/2} (x_{[2i-1]} \hspace{-0.8mm}+\hspace{-0.8mm} x_{[2i]}^2\hspace{-0.8mm}-\hspace{-0.8mm}1)^2 \end{array}$  }  \hspace{-4.5mm} &  \hspace{-4.5mm}  \scriptsize{ $\begin{array}{c} n=100 \\ \textrm{[1,0.1,$\ldots$, 1.1, 0.1$]^T$}  \end{array}$ }  \hspace{-5mm} 
\\ \hline
 \hspace{-6mm} \scriptsize{ $\begin{array}{c}\textrm{fun3:} \\ \textrm{Quadratic QF1:} \end{array}$} \hspace{-4.5mm} & \hspace{-4.5mm}  {\scriptsize $\begin{array}{l} \frac{1}{2}\sum_{i=1}^{n}i x_{[i]}^2 -x_{[n]}   \end{array}$  }  \hspace{-4.5mm} &  \hspace{-4.5mm}  \scriptsize{ $\begin{array}{c} n=100 \\ \textrm{[1,1,$\ldots$, 1$]^T$}  \end{array}$ }  \hspace{-5mm} 
\\ \hline
 \hspace{-6mm} \scriptsize{ $\begin{array}{c}\textrm{fun4:} \\ \textrm{Extended BD1:} \end{array}$} \hspace{-4.5mm} & \hspace{-4.5mm}  {\scriptsize $\begin{array}{l} \sum_{i=1}^{n/2}(x_{[2i-1]}^2+x_{[2i]}^2-2)  \\ +\hspace{-0.8mm} \sum_{i=1}^{n/2} (\textrm{exp}(x_{[2i-1]}) \hspace{-0.8mm} -\hspace{-0.8mm} x_{[2i]})^2  \end{array}$  }  \hspace{-4.5mm} &  \hspace{-4.5mm}  \scriptsize{ $\begin{array}{c} n=100 \\ \textrm{[0.1,0.1,$\ldots$, 0.1$]^T$}  \end{array}$ }  \hspace{-5mm} 
\\ \hline
 \hspace{-6mm} \scriptsize{ $\begin{array}{c}\textrm{fun5:} \\ \textrm{TRIDIA:} \end{array}$} \hspace{-4.5mm} & \hspace{-4.5mm}  {\scriptsize $\begin{array}{l} (x_{[1]}-1)^2 \\ +\sum_{i=2}^n i (2x_{[i]}-x_{[i-1]})^2  \end{array}$  }  \hspace{-4.5mm} &  \hspace{-4.5mm}  \scriptsize{ $\begin{array}{c} n=100 \\ \textrm{[1,1,$\ldots$, 1$]^T$}  \end{array}$ }  \hspace{-5mm} 
\\ \hline
 \hspace{-6mm} \scriptsize{ $\begin{array}{c}\textrm{fun6:} \\ \textrm{ARWHEAD:} \end{array}$} \hspace{-4.5mm} & \hspace{-4.5mm}  {\scriptsize $\begin{array}{l} \sum_{i=1}^{n-1}(-4x_{i}+3) \\+\sum_{i=1}^{n-1} (x_{[i]}^2+x_{[n]}^2)^2  \end{array}$  }  \hspace{-4.5mm} &  \hspace{-4.5mm}  \scriptsize{ $\begin{array}{c} n=100 \\ \textrm{[1,1,$\ldots$, 1$]^T$}  \end{array}$ }  \hspace{-5mm} 
\\ \hline
 \hspace{-6mm} \scriptsize{ $\begin{array}{c}\textrm{fun7:} \\ \textrm{EG2:} \end{array}$} \hspace{-4.5mm} & \hspace{-4.5mm}  {\scriptsize $\begin{array}{l} \sum_{i=1}^{n-1}\textrm{sin}(x_{[1]}+x_{[i]}^2-1) \\ +\frac{1}{2}\textrm{sin}(x_{[n]}^2)  \end{array}$  }  \hspace{-4.5mm} &  \hspace{-4.5mm}  \scriptsize{ $\begin{array}{c} n=100 \\ \textrm{[1,1,$\ldots$, 1$]^T$}  \end{array}$ }  \hspace{-5mm} 
\\ \hline
 \hspace{-5.5mm} \scriptsize{ $\begin{array}{c}\textrm{fun8: Partial} \\ \textrm{ perturbed quadratic } \end{array}$} \hspace{-4.5mm} & \hspace{-4.5mm}  {\scriptsize $\begin{array}{l} x_{[1]}^2 + \sum_{i=1}^{n} ix_{[i]}^2 \\ +\sum_{i=1}^n\frac{1}{100}(x_{[1]}\hspace{-0.8mm}+\ldots+\hspace{-0.8mm}x_{[i]})^2  \end{array}$  }  \hspace{-4.5mm} &  \hspace{-4.5mm}  \scriptsize{ $\begin{array}{c} n=100 \\ \textrm{[1,1,$\ldots$, 1$]^T$}  \end{array}$ }  \hspace{-5mm} 
\\ \hline
 \hspace{-6mm} \scriptsize{ $\begin{array}{c}\textrm{fun9: ENGVAL1}  \end{array}$} \hspace{-4.5mm} & \hspace{-4.5mm}  {\scriptsize $\begin{array}{l} 
 \sum_{i=1}^{n-1} (x_{[i]}^2+x_{[i+1]}^2)^2 \\ +\sum_{i=1}^{n-1}(-4x_{[i]}+3)  \end{array}$  }  \hspace{-4.5mm} &  \hspace{-4.5mm}  \scriptsize{ $\begin{array}{c} n=100 \\ \textrm{[2,2,$\ldots$, 2$]^T$}  \end{array}$ }  \hspace{-5mm} 
\\ \hline
 \hspace{-6mm} \scriptsize{ $\begin{array}{c}\textrm{fun10: EDENSCH}  \end{array}$} \hspace{-4.5mm} & \hspace{-4.5mm}  {\scriptsize $\begin{array}{l} 
 16\hspace{-0.8mm}+\hspace{-0.8mm}\sum_{i=1}^{n-1}\hspace{-0.8mm}\Big[ (x_{[i]}\hspace{-0.8mm}-\hspace{-0.8mm}2)^4\hspace{-0.8mm}+\hspace{-0.8mm}(x_{[i]}x_{[i+1]} \\ -2x_{[i+1]})^2\hspace{-0.8mm}+\hspace{-0.8mm}(x_{[i+1]}\hspace{-0.8mm}+\hspace{-0.8mm}1)^2 \Big] \end{array}$  }  \hspace{-4.5mm} &  \hspace{-4.5mm}  \scriptsize{ $\begin{array}{c} n=100 \\ \textrm{[0,0,$\ldots$, 0$]^T$}  \end{array}$ }  \hspace{-5mm} 
\\ \hline
\end{tabular}
\vspace{0mm}
\end{table}

\begin{figure}[t!]
\centering
\includegraphics[width=80mm]{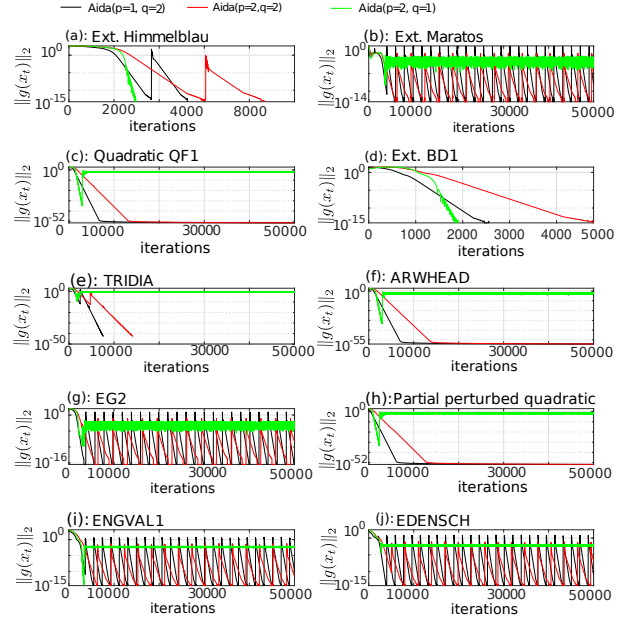}
\caption{\small  Evaluation of Aida for solving ten optimisation problems (see Table~\ref{tab:toy}). The parameters of Aida except $(p,q)$ are set to be $(\beta_1, \beta_2, \mu, \eta, \epsilon)= (0.9, 0.99, 0, 0.001, 1e-50)$. } 
\label{fig:Aida_toy}
\vspace{-2mm}
\end{figure}

The convergences results are summarised in Fig.~\ref{fig:Aida_toy} with respect to gradient norm $\|g(x_t)\|_2$ versus iterations, where one subplot for each problem.  It is seen that the behaviour of Aida changes across different problems. In Fig.~\ref{fig:Aida_toy}:$\{(a), (d)\}$, the setup $(p,q)=(2,1)$ outperforms $(p,q)=(2,2)$ and $(p,q)=(1,2)$, indicating that AdamW may beat Aida with $q>1$ for certain optimisation problems.  On the other hand, in Fig.~\ref{fig:Aida_toy}:$\{(c), (e), (f), (h)\}$, the setups $(p,q)=(2,2)$ and $(p,q)=(1,2)$  produce better performance than $(p,q)=(2,1)$.

Considering Fig.~\ref{fig:Aida_toy}:$\{(b), (g), (i), (g)\}$, one observes that Aida with the setups $(p,q)=(2,2)$ and  $(p,q)=(1,2)$ exhibit periodic patterns while the setup $(p,q)=(2,1)$ does not. The above results suggest that $q>1$ makes Aida to repeatedly visit some satisfactory suboptimal solutions, which is more desirable than the behaviour of  $(p,q)=(2,1)$ of which the curves fluctuate around certain unsatisfactory suboptimal solutions.   

Finally, it is clear from the ten plots that the setup $(p,q)=(1,2)$ converges faster than $(p,q)=(2,2)$, which is consistent with the local convergence results of Fig.~\ref{fig:Aida_quad_beta_eta}. This suggest that in practice, it might be preferable to select $(p,q)=(1,2)$ than $(p,q)=(2,2)$.

\begin{remark}
We emphasise that that development of Aida with $q>1$ in this paper is not to replace AdamW (i.e., Aida with $(p,q)=(2,1)$) in all optimisation problems. In practice, one should  set the proper values for $p$ and $q$ in Aida depending on the particular applications. 
\end{remark}

\subsection{Evaluation of Aida via DNN training}
\noindent\textbf{Training of Transformer for an NLP task}: We apply Aida in training a Transformer for WMT16: multimodal translation.  The two setups $(p,q)=(2,1)$ and $(p,q)=(1,2)$ were tested in Aida. To make a fair comparison, we adopted an existing open-source,\footnote{https://github.com/jadore801120/attention-is-all-you-need-pytorch} which produces reasonable validation performance using Adam. 

In the training process, we retained almost all of the default hyper-parameters provided in the open-source except the batchsize. Due to limited GPU memory, we changed the batchsize from 256 to 200. The default hyper-parameters are $(\beta_1, \beta_2)=(0.9, 0.98)$ and $\epsilon=1e-9$. As indicated in the open-source, no weight decay was imposed in either AdamW or Aida in our experiments (i.e., $\mu=0$).  To alleviate the effect of the randomness in the training process, three experimental repetitions were conducted for each setup of Aida.  Our primary interest is the validation performance gain of $(p,q)=(1,2)$ over $(p,q)=(2,1)$.  


 \begin{figure}[t!]
\centering
\includegraphics[width=80mm]{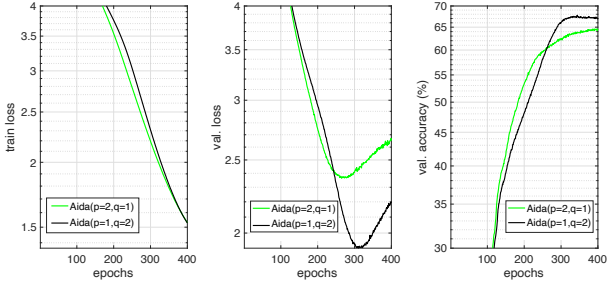}
\caption{Performance comparison of two setups of Aida for training a Transformer over WMT16:multimodal translation. The curve for each setup is obtained by averaging the results from three experimental repetitions. It is noted that the setup $(p,q)=(2,1)$ of Aida is equivalent to Adam.  } 
\label{fig:Aida}
\end{figure}

\begin{table}[t]
\caption{\small  Validation accuracy for training  a Transformer   \vspace{2mm} } 
\label{tab:transformer_val_acc}
\centering
\begin{tabular}{|c|c|c|}
\hline
& \hspace{0mm} {{\scriptsize Aida(p=1,q=2)}}\hspace{0mm}
& \hspace{0mm}{\scriptsize  $\begin{array}{c}\textrm{Aida(p=2,q=1)} \\ (\textrm{equivalent to Adam}) \end{array}$ }   \hspace{0mm} 
 \\
\hline
\hline  
\footnotesize{val. acc.} & \footnotesize{\textbf{67.8$\pm$ 0.28} } & \footnotesize{64.76$\pm$ 0.63} \\ 
\hline
\end{tabular}
\vspace{-3mm}
\end{table}

 Fig.~\ref{fig:Aida} displays the convergence results of the two setups of Aida. Each curve in the plot is obtained by averaging the convergence results from the three experimental repetitions. It is seen that the convergence behaviours of the training loss are quite similar for $(p,q)=(2,1)$ and $(p,q)=(1,2)$. When it comes to validation loss and accuracy, the setup $(p,q)=(1,2)$ produces considerably better convergence results than $(p,q)=(2,1)$. The performance impact by changing from $(p,q)=(2,1)$ to $(p,q)=(1,2)$ in Aida is significant.  Table \ref{tab:transformer_val_acc} shows the averaged validation accuracy of three experimental repetitions per setup of Aida.  The performance gain of  $(p,q)=(1,2)$ is roughly 3\% over the setup $(p,q)=(2,1)$, suggesting the effectiveness of  $(p,q)=(1,2)$. 


\noindent\textbf{Training of Swin-Transformer over CIFAR10}: The recent work  \cite{liu2021Swin} proposes Swin-Transformer, a variant of Transformer,  for vision applications. It is reported in \cite{liu2021Swin} that the new DNN model produces state-of-the-art results in few vision tasks. In this work, we evaluate Aida for training Swin-Transformer over CIFAR10.  Similar to the NLP task, the two setups $(p,q)=(2,1)$ and $(p,q)=(1,2)$ for Aida were investigated. Two values of $\beta_2$ were tested, which are 0.98 and 0.999.  The other parameters were set to $(\beta_1, \mu, \epsilon)=(0.9, 0, 1e-9)$. The parameters for the training procedure and Swin-Transformer were summarised in Appendix~\ref{appendix:swin-Transformer_param}. Similarly, three experimental repetitions were conducted for each setup of Aida.

Table~\ref{tab:swin_val_acc} summarises the validation accuracies. It is clear that Aida with $(p,q)=(1,2)$ outperforms $(p,q)=(2,1)$ with an accuracy gain of around 3\%. The above results indicate that for Aida with $q>1$ produces promising performance not only in NLP tasks but also in vision tasks when training Transformer and its variant. 
  
\begin{table}[t]
\caption{\small Validation accuracy for training Swin-Transformer  \vspace{2mm} } 
\label{tab:swin_val_acc}
\centering
\begin{tabular}{|c|c|c|c|}
\hline
\hspace{-4.5mm} & \hspace{-4.5mm} {{\scriptsize $\begin{array}{c}\textrm{Aida(p=1,q=2)} \\ ({\beta_2=0.98}) \end{array}$ }} \hspace{0mm}
\hspace{-4.5mm} & \hspace{-4.5mm} {\scriptsize  $\begin{array}{c}\textrm{Aida(p=2,q=1)} \\ ({\beta_2=0.98}) \end{array}$ }   \hspace{0mm} 
\hspace{-4.5mm} & \hspace{-4.5mm} {\scriptsize  $\begin{array}{c}\textrm{Aida(p=2,q=1)} \\ ({\beta_2=0.999}) \end{array}$ }   \hspace{-4.5mm} 
 \\
\hline
\hline  
\footnotesize{val. acc.} & \footnotesize{\textbf{89.25}$\pm$0.29} & \footnotesize{85.86$\pm$0.23} &   \footnotesize{78.23$\pm$2.52}  \\ 
\hline
\end{tabular}
\vspace{0mm}
\end{table}

\vspace{-2mm}
\section{Conclusions}
\label{sec:conclusion}
\vspace{-1mm}

In this work, we have proposed a new adaptive optimisation method named Aida as an extension of AdamW.  Our motivation for developing Aida was to relax the local convergence condition for Adam analysed in \cite{Bock19Adam}, which requires the learning rate to be sufficiently small to ensure local stability around an optimal solution of a twice-differentiable function. Aida is designed to generalise the update expressions of AdamW in two aspects. Firstly, the 2nd moment $\{r_t|t>0\}$ of the $p$th power of the gradient magnitudes are tracked as an extension of the 2nd moment $\{v_t |t>0 \}$ of squared-gradients in AdamW. Secondly,  the so-called magnitude vector of the update direction in Aida is defined as $ \frac{ |{m}_{t+1}|^q }{ ( r_{t+1}^{1/p}  \hspace{-0.mm} + \hspace{-0.mm}  \epsilon)^q}$ $\left(\textrm{or } \frac{ |{m}_{t+1}|^q }{ r_{t+1}^{q/p}  \hspace{-0.mm} + \hspace{-0.mm}  \epsilon}\right)$ instead of  $ \frac{|{m}_{t+1}| }{\sqrt{v_{t+1} \hspace{-0.mm} + \hspace{-0.mm}  \epsilon} }$ $\left(\textrm{or } \frac{|{m}_{t+1}| }{(\sqrt{v_{t+1}} \hspace{-0.mm} + \hspace{-0.mm}  \epsilon}\right)$ in AdamW.  The parameter $q$ in Aida is introduced to control the impact of the element-wise learning rates. When $(p,q)=(2,1)$, Aida reduces to AdamW. 

Both theoretical analysis and empirical study have been conducted for Aida. It is shown that when $p>1$, $q>1$, an optimal solution for a  twice-differentiable function is locally stable only when the weight decay is non-zero. There is no explicit requirement on learning rate being sufficiently small. We further investigate the global convergence (i.e., the initial value $x_0$ is not close to $x_{\ast}$) of Aida for solving ten toy optimisation problems, which suggests that for a number of problems, Aida with $(p,q)=(1,2)$ performs better than $(p,q)=(2,1)$. Aida is also evaluated for training Transformer and its variant for two deep learning tasks. Experimental results indicate that Aida with $(p,q)=(1,2)$ outperforms $(p,q)=(2,1)$ by a significant gain in terms of validation accuracy. The above results suggest the parameter $(p,q)$ in Aida should be properly configured depending on the particular optimisation problem to be minimised.


\appendices 

\section{Setups for training Swin-Transformer over CIFAR10} 
\label{appendix:swin-Transformer_param}

\begin{table}[h]
\caption{\small Parameter setups for the training procedure} 
\label{tab:swin_param}
\centering
\begin{tabular}{|c|c|}
\hline
{\footnotesize batchsize} & {\footnotesize 128} \\
\hline
{\footnotesize epochs} & {\footnotesize 200} \\
\hline
{\footnotesize iteration steps per epoch} & {\footnotesize 391} \\
\hline
{\footnotesize learning rate scheduling} & {\footnotesize CosineLRScheduler} \\
\hline
{\footnotesize warmup steps} & {\footnotesize 5474} \\
\hline
\end{tabular}
\vspace{0mm}
\end{table}

\begin{table}[h]
\caption{\small Parameter setups for swin-transformer} 
\label{tab:swin_param}
\centering
\begin{tabular}{|c|c|}
\hline
{\footnotesize patch size} & {\footnotesize 2} \\
\hline
{\footnotesize embed dimension } & {\footnotesize 96} \\
\hline
{\footnotesize depths} & {\footnotesize $[2,2,6, 2]$} \\
\hline
{\footnotesize number of heads} & {\footnotesize $[3, 6, 12, 24]$} \\
\hline
{\footnotesize window size} & {\footnotesize $2$} \\
\hline
{\footnotesize MLP ratio} & {\footnotesize $10$} \\
\hline
{\footnotesize drop path rate} & {\footnotesize $0.1$} \\
\hline
{\footnotesize drop rate} & {\footnotesize $0.0$} \\
\hline
\end{tabular}
\vspace{0mm}
\end{table}

%
\IEEEpeerreviewmaketitle

\ifCLASSOPTIONcaptionsoff
  \newpage
\fi


\begin{thebibliography}{10}
\providecommand{\url}[1]{#1}
\csname url@samestyle\endcsname
\providecommand{\newblock}{\relax}
\providecommand{\bibinfo}[2]{#2}
\providecommand{\BIBentrySTDinterwordspacing}{\spaceskip=0pt\relax}
\providecommand{\BIBentryALTinterwordstretchfactor}{4}
\providecommand{\BIBentryALTinterwordspacing}{\spaceskip=\fontdimen2\font plus
\BIBentryALTinterwordstretchfactor\fontdimen3\font minus
  \fontdimen4\font\relax}
\providecommand{\BIBforeignlanguage}[2]{{%
\expandafter\ifx\csname l@#1\endcsname\relax
\typeout{** WARNING: IEEEtran.bst: No hyphenation pattern has been}%
\typeout{** loaded for the language `#1'. Using the pattern for}%
\typeout{** the default language instead.}%
\else
\language=\csname l@#1\endcsname
\fi
#2}}
\providecommand{\BIBdecl}{\relax}
\BIBdecl

\bibitem{Andrei08optFamlily}
N. Andrei.
\newblock {An Unconstrained Optimization Test Functions Collection}.
\newblock {\em Advanced Modeling and Optimization}, 10:147--161, 2008.

\bibitem{Balles18Adam}
L. Balles and P. Hennig.
\newblock {Dissecting Adam: The Sign, Magnitude and Variance of Stochastic
  Gradients}.
\newblock In {\em ICML}, 2018.

\bibitem{Bock18Adam}
S. Bock, J. Goppold, and M. Weib.
\newblock {An improvement of the convergence proof of the ADAM-Optimizer}.
\newblock arXiv:1804.10587 [cs.LG], 2018.

\bibitem{Bock19Adam}
S. Bock and M. Weib.
\newblock {A Proof of Local Convergence for the AdamOptimizer}.
\newblock In {\em International Joint Conference on Neural Networks (IJCNN)},
  2019.

\bibitem{Bof18dynamic}
N. Bof, R. Carli, and L. Schenato.
\newblock {Lyapunov Theory for Discrete Time Systems}.
\newblock arXiv:1809.05289 [math.OC], 2018.

\bibitem{Dozat16NAdam}
T. Dozat.
\newblock {Incorporating Nesterov Momentum into Adam}.
\newblock In {\em International conference on Learning Representations (ICLR)},
  2016.

\bibitem{Duchi11AdaGrad}
J. Duchi, E. Hazan, and Y. Singer.
\newblock {Adaptive Subgradient Methods for Online Learning and Stochastic
  Optimization}.
\newblock {\em Journal of Machine Learning Research}, 12:2121--2159, 2011.

\bibitem{Gemp19Adam}
I. Gemp and B. McWilliams.
\newblock {The Unreasonable Effectiveness of Adam on Cycles}.
\newblock In {\em NeurIPS}, 2019.

\bibitem{Goodfellow14GAN}
I. Goodfellow, J. Pouget-Abadie, M. Mirza, B. Xu, D. Warde-Farley, S. Ozair, A.
  Courville, and Y. Bengio.
\newblock {Generative Adversarial Nets}.
\newblock In {\em Proceedings of the International Conference on Neural
  Information Processing Systems}, pages 2672--2680, 2014.

\bibitem{He15ResNet}
K. He, X. Zhang, S. Ren, and J. Sun.
\newblock {Deep Residual Learning for Image Recognition}.
\newblock In {\em IEEE conference on Computer Vision and Pattern Recognition
  (CVPR)}, 2015.

\bibitem{Devlin18Bert}
K.~Lee J.~Devlin, M-W.~Chang and K. Toutanova.
\newblock {Bert: Pre-training of deep bidirectional transformers for language
  understanding}.
\newblock arXiv:1810.04805, 2018.

\bibitem{Zhang2019Adam}
A.~Veit S. Kim S. J. Reddi S. Kumar S.~Sra J.~Zhang, S. P.~Karimireddy.
\newblock {Why ADAM Beats SGD for Attention Models}.
\newblock In {\em submitted for review by ICLR}, 2019.

\bibitem{Karimi16PLcon}
H. Karimi, J. Nutini, and M. Schmidt.
\newblock {Linear Convergence of Gradient and Proximal-Gradient Methods Under
  the Polyak-Lojasiewicz Condition}.
\newblock arXiv:1608.04636 [cs.LG], 2016.

\bibitem{Kingma17}
D.~P. Kingma and J.~L. Ba.
\newblock {Adam: A Method for Stochastic Optimization}.
\newblock arXiv preprint arXiv:1412.6980v9, 2017.

\bibitem{Lecun15nature}
Y. LeCun, Y. Bengio, and G. Hinton.
\newblock {Deep Learning}.
\newblock {\em Nature}, 521:436--444, 2015.

\bibitem{liu2021Swin}
Ze Liu, Yutong Lin, Yue Cao, Han Hu, Yixuan Wei, Zheng Zhang, Stephen Lin, and
  Baining Guo.
\newblock {Swin Transformer: Hierarchical Vision Transformer using Shifted
  Windows}.
\newblock In {\em International Conference on Computer Vision (ICCV)}, 2021.

\bibitem{Loshchilov19AdamW}
I. Loshchilov and F. Hutter.
\newblock {Decoupled Weight Decay Regularization}.
\newblock In {\em ICLR}, 2019.

\bibitem{Polyak64CM}
B.~T. Polyak.
\newblock {Some methods of speeding up the convergence of iteration methods}.
\newblock {\em USSR Computational Mathematics and Mathematical Physics},
  4:1--17, 1964.

\bibitem{Riedmiller93signOpt}
M. Riedmiller and H. Braun.
\newblock {A direct adaptive method for faster backpropagation learning: the
  RPROP algorithm}.
\newblock In {\em IEEE International Conference on Neural Networks}, 1993.

\bibitem{Rubio17Adam}
D.~M. Rubio.
\newblock {Convergence Analysis of an Adaptive method of Gradient Descent}.
\newblock Master's thesis, University of Oxford, 2017.

\bibitem{Reddi18Amsgrad}
S.~Kale S.~J.~Reddi and S. Kumar.
\newblock {On the Convergence of Adam and Beyond}.
\newblock In {\em International conference on Learning Representations (ICLR)},
  2018.

\bibitem{Seide14signOpt}
F. Seide, H. Fu, J. Droppo, G. Li, and D. Yu.
\newblock {1-bit stochastic gradient descent and its application to
  data-parallel distributed training of speech DNNs}.
\newblock In {\em Fifteenth Annual Conference of the International Speech
  Communication Asssociation}, 2014.

\bibitem{Shazeer18Adafactor}
N. Shazeer and M. Stern.
\newblock {Adafactor: Adaptive Learning Rates with Sublinear Memory Cost}.
\newblock arXiv:1804.04235 [cs.LG], 2018.

\bibitem{Sutskever13NAG}
H. Sutskever, J. Martens, G. Dahl, and G. Hinton.
\newblock {On the importance of initialisation and momentum in deep learning}.
\newblock In {\em International conference on Machine Learning (ICML)}, 2013.

\bibitem{Tieleman12RMSProp}
T. Tieleman and G. Hinton.
\newblock {Lecture 6.5-RMSProp: Divide The Gradient by a Running Average of Its
  Recent Magnitude}.
\newblock COURSERA: Neural networks for machine learning, 2012.

\bibitem{Transformer17}
A. Vaswani, N. Shazeer, N. Parmar, J. Uszkoreit, L. Jones, A.~N. Gomez, L.
  Kaiser, and I. Polosukhin.
\newblock Attention is all you need.
\newblock arXiv:1706.03762 [cs. CL], 2017.

\bibitem{Wilson17AdamNegative}
A.~C. Wilson, R. Roelofs, M. Stern, N. Srebro, and B. Recht.
\newblock {The Marginal Value of Adaptive Gradient Methods in Machine
  Learning}.
\newblock In {\em 31st Conference on Neural Information Processing Systems
  (NIPS)}, 2017.

\bibitem{You20Lamb}
Y. You, J. Li, S. Reddi, J. Hseu, S. Kumar, S. Bhojanapalli, X. Song, J.
  Demmel, K. Keutzer, and C.-J. Hsieh.
\newblock {Large Batch Optimization for Deep Learning: Training BERT in 76
  minutes}.
\newblock In {\em ICRL}, 2020.

\bibitem{Zhang20AdamBetter}
J. Zhang, S.~P. Karimireddy, A. Veit, S. Kim, S. Reddi, and S. Sra.
\newblock {Why Adam Beats SGD for Attention Models}.
\newblock Under review by ICLR, 2020.


\end{thebibliography}
\end{document}